# Cluster-based Zero-shot learning for multivariate data

Toshitaka Hayashi [a, 1] and Hamido Fujita [a, 2][0000-0001-5256-210X]

[a] *Faculty of Software and Information Science, Iwate Prefectural University, Iwate, Japan*

[1]G236r002@s.iwate-pu.ac.jp

[2]issam@iwate-pu.ac.jp

**Abstract.** Supervised learning requires a sufficient training dataset which includes all label. However, there are cases that some class is not in the training data. Zero-Shot Learning (ZSL) is the task of predicting class that is not in the training data (target class). The existing ZSL method is done for image data. However, the zero-shot problem should happen to every data type. Hence, considering ZSL for other data types is required. In this paper, we propose the cluster-based ZSL method, which is a baseline method for multivariate binary classification problems. The proposed method is based on the assumption that if data is far from training data, the data is considered as target class. In training, clustering is done for training data. In prediction, the data is determined belonging to a cluster or not. If data does not belong to a cluster, the data is predicted as target class. The proposed method is evaluated and demonstrated using the KEEL dataset.

## 1 Introduction

General supervised learning algorithms require training data for all labels. However, labeling data is a difficult task. Hence, there is a case that data for some labels is not in the training data. In this case, supervised learning algorithms is not suitable. Zero-Shot Learning (ZSL, Also known as zero-data learning) (Larochelle et al. 2008) is developed to solve the problem. The goal of ZSL is predicting unknown labels, which are not in the training data. ZSL could be applied to detect non-labeled data and can reduce the task of data labeling.

In ZSL, there are two types of class, Train-Class and Target-Class. Train-Class is in the training data. Target-Class is not in the training data. Hence, ZSL can train only data on Train-Class, but it has to predict train-class and target-class.

Most ZSL research is done on image data (Wang et al. 2019). However, the zero-shot problem should happen not only image data but also every data type. Hence considering ZSL for other data types is an important problem. Also, every data type can be represented by a vector. Hence considering ZSL on vector space is the most important problem. It is because if ZSL works on vector space, ZSL works on every data type. In this paper, we consider ZSL for vector space. In ZSL, it could consider data that is not in training data is the target class. Hence, the prediction could be made based on the assumption data that is far from the training data is the Target Class. Based on the above, we proposed the cluster-based ZSL algorithm. K-cluster is created from training data, and prediction is made using the distance between data and centroids of the clusters.

## 2 Related Work

The supervised classification has achieved much success in many areas. However, in supervised learning, sufficient labeled training data is required for each class. Also, the classifier can only predict the class that is in training data and cannot predict the class that is not in training data (Wang et al. 2019).

For example, the classification problem for data marked as A and B is shown in Fig. 1. Fig.1 shows well-labeled data and zero-shot data. In well-labeled data, traditional classification algorithms can find boundary. However, traditional classification algorithms are not suitable for zero-shot data because finding a boundary on something that does not have data is difficult. In order to classify zero-shot data, different classification algorithms are required.

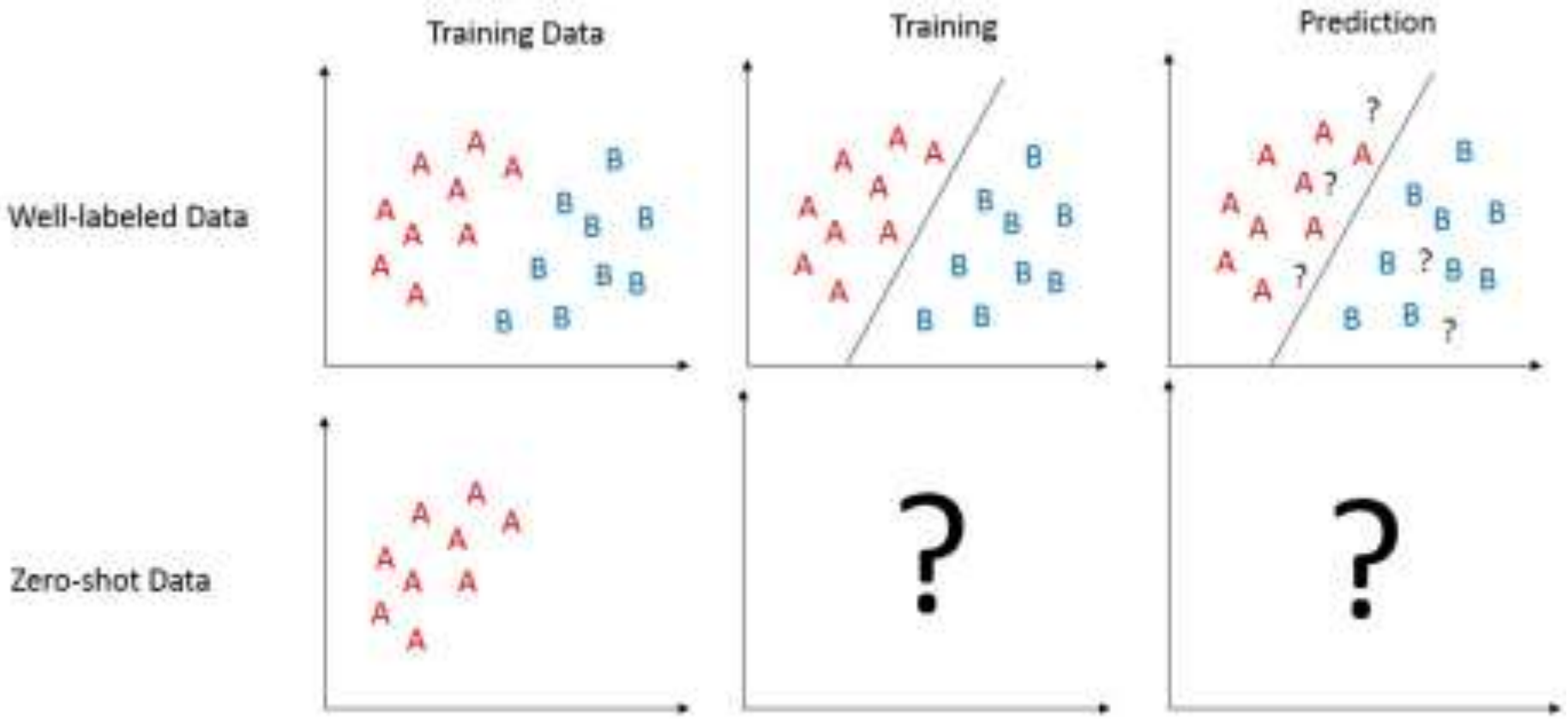

**Fig. 1.** Zero-shot problem for supervised learning

The goal of ZSL is to predict class that is not in training data. In ZSL, there are some labeled training data. The class covered by this training data is referred to as train(seen) class. Also, there are some unlabeled testing data, whose class is not covered by training data. The class is referred to as a target(unseen) class. The initial problem of ZSL is training only train class. Also, test data includes only target class. Hence ZSL is considered as a subfield of transfer learning(Pan et al. 2010,Wang et al. 2019). The model is trained by training class and applied to predict the target class. However, some researches pointed out that the initial problem is unrealistic(Xian et al 2017, Wang et al. 2019); test data should include train class and target class. Hence, Generalized Zero-shot Learning is introduced(Socher et al. 2013, Chao et al. 2016, Xian et al. 2017, Liu et al 2018,Wang et al. 2019). In generalized zero-shot learning, test data includes both train class and target class. Also, recognition of the target class is required.

The most popular ZSL methods are using compatibility function between image feature space and label space (Larochelle et al.2008, Frome et al. 2013, Akata et al. 2013, Akata et al. 2015, Paredes et al. 2015). In order to represent label vector space, word2vec(Mikolov et al. 2013) is utilized. On other hand, Socher et al. (2013) utilized outlier detection to recognize target class. In their method, if data is an outlier, the data is predicted as the target class. Socher et al. (2013) proposed two type outlier detection strategies, using Gaussian model and Local Outlier Probability model (Kriegel et al. 2009).

Moreover, the most existing ZSL research has been done mainly for image data (Larochelle et al.2008, Socher et al. 2013, Frome et al. 2013, Akata et al. 2013, Akata et al. 2015, Paredes et al. 2015, Chao et al. 2016, Xian et al. 2017, Liu et al 2018,Wang et al. 2019). However, the zero-shot problem shoul happen not only image data but also other kind of data type.In this paper, we consider ZSL for multivariate data because every data type could be represented as a vector. Also, applying ZSL algorithm for image data to multivariate data is not suitable. This is because ZSL methods for image data requires semantic label information (Socher et al. 2013, Frome et al. 2013, Akata et al. 2013, Akata et al. 2015, Paredes et al. 2015, Chao et al. 2016, Xian et al. 2017, Liu et al 2018, Wang et al. 2019) while many multivariate data does not have such information (Fdez et al. 2011). Hence new approach is necessary.

In this paper, we consider ZSL for multivariate binary classification problem. The main problem is how to train data and how to predict the corresponding target class.

## 3 Proposed method

In this section, we propose a cluster-based ZSL method. The goal is predicting train class or target class. In zero-shot learning, only train class is available for training. Hence, assumption is needed to predict target class. The proposed method is based on the assumption that data for target class is distant from training data. In order to determine whether data is distant, representing training data and calculating distance are necessary. On the basis of above, we propose cluster-based zero-shot learning method as shown in Fig.2. In our proposal, clustering is done to represent training data. Also, distance is calculated to determine whether data is distant.

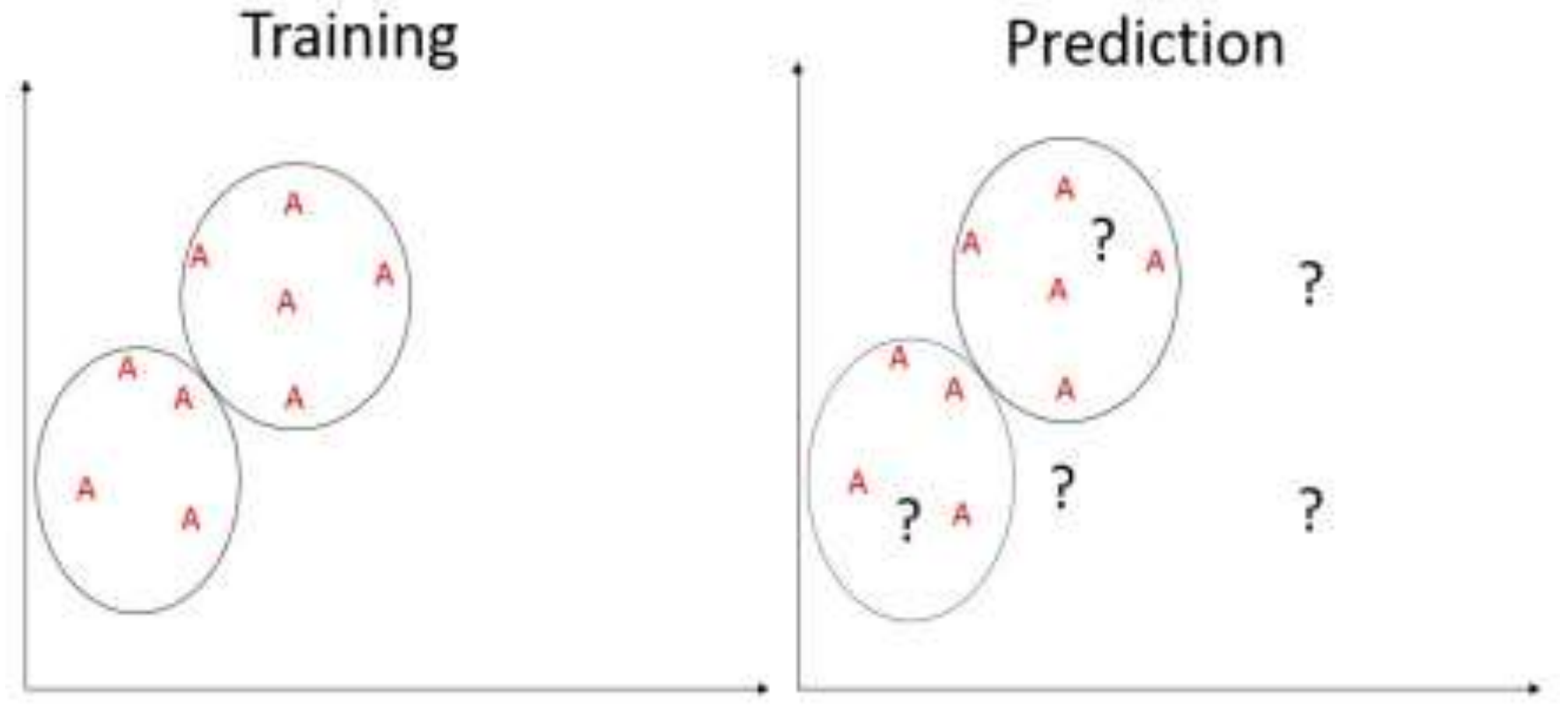

**Fig. 2.** Cluster-based Zero-shot learning

Proposed method consists of two steps, training and prediction.
In training, create k-clusters from the training dataset. The training dataset is constructed by only data for the train class. Hence, all clusters are the representation of train class. Then we calculate thresholds which are the radius of the clusters. If the distance between the data and the cluster is smaller than the threshold, the data belongs to the cluster.

In prediction, we check if the data belongs to any cluster or not. Whether it belongs to a cluster is determined by the distance between the data and the nearest centroid. If data does not belong to any cluster, the data is treated as Target Class. Otherwise, the data is treated as Train Class.

On the basis of the above, we propose cluster-based zero-shot learning framework as shown in Fig 3. K-Clusters and K-thresholds are extracted from training data. Prediction is done by checking data belonging to cluster or not. More details about training and prediction are shown in section 3.1 and section 3.2.

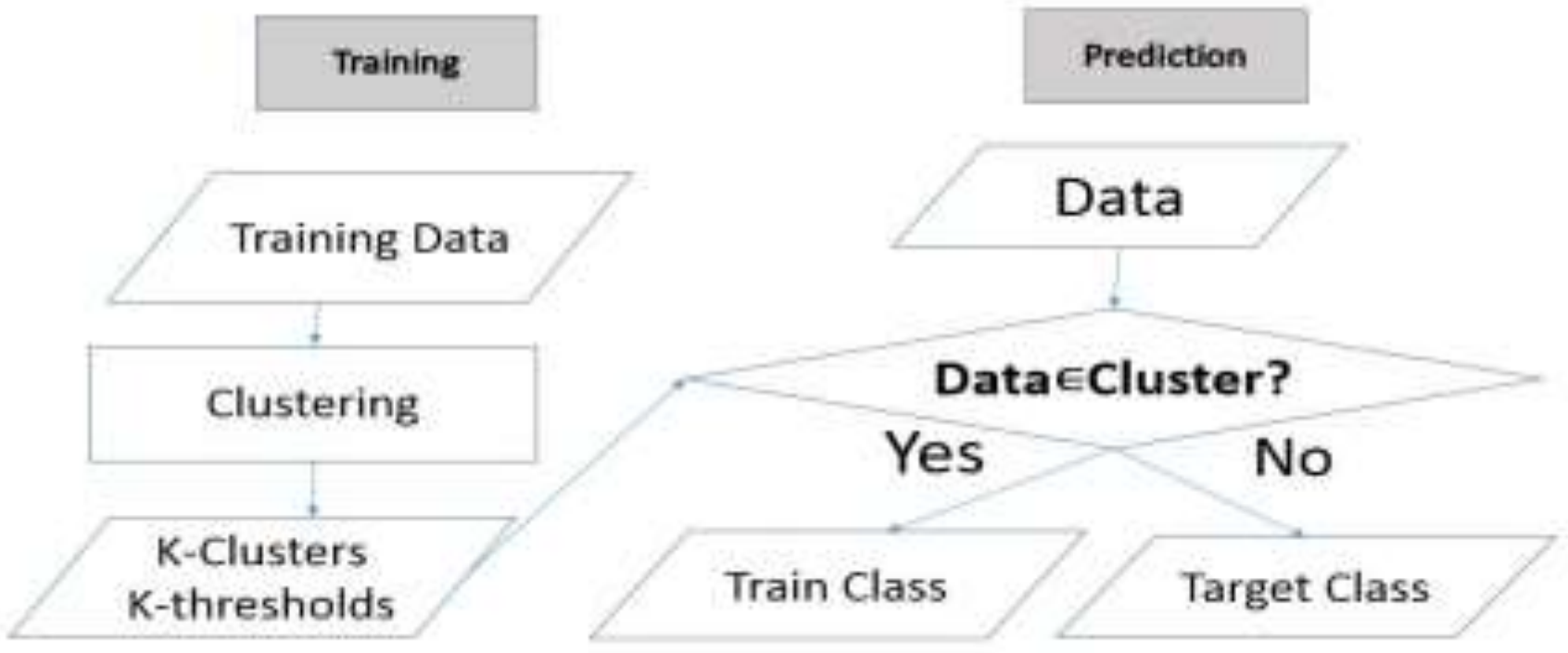

**Fig. 3.** Cluster-based Zero-shot learning framework

### 3.1 Training

In training, K-Clusters (Centroids) and K-thresholds are extracted from training data. In this paper, we use k-means for simplicity. Training is done as Fig.4. In Fig.4, K is the parameter. K is determined as the experiment.

**Input**: Training Data (Train), K
**Output**: K-Centroids ($C_1$, $C_2$,..., $C_k$), K-Thresholds ($T_1$,$T_2$,...,$T_k$)
1. Make K-Cluster from training dataset.
2. Calculate thresholds for each cluster. The model should treat all training data as train class. Hence, for each cluster, threshold is the distance between centroid and farthest data (data belongs to cluster) from the centroid. Threshold $T_k$ is extracted as given in formula (1). Centroid $C_k$ is the average of the data in the $\text{Cluster}_k$. Hence $C_k$ is calculated as given in formula (2). In order to calculate distance, Euclidean distance is utilized. The formula is as given in formula (3).

**Fig. 4.** Process of Training

$$T_K = \{\max\left(distance(data, C_K)\right) \,|\, data \in (Train \cap Cluster_K)\} \quad (1)$$

$$C_k = \{\frac{\sum data}{|\text{data}|} \,|\, data \in (Train \cap Cluster_K)\} \quad (2)$$

$$distance(data, C) = \sqrt{\sum_{d=1}^{n} (data_d - C_d)^2} \quad (3)$$

### 3.2 Prediction

The goal of prediction is the classification of train class and target class. If data is far from the nearest cluster, the data is considered as target class. Which means if distance between data and centroid of the nearest cluster is higher than threshold, the data is treated as target class. Hence prediction is done as Fig.5.

**Input**: Test data
**Output**: Prediction
$C_n$ is the nearest centroid from test data.

IF distance (data, $C_n$)>threshold $T_n$:
$$\text{Prediction}(\text{data}) = \text{Target class}$$
ELSE:
$$\text{Prediction}(\text{data}) = \text{Train class}$$

**Fig. 5.** Process of Prediction

# 4 Result

The proposed approach has been validated against data listed in section 4.1. The measurement of evaluation is shown in sectoin 4.2. The experiment results are shown in section 4.3.

## 4.1 The data

In this paper, we use KEEL datasets(Fdez et al.2011) for the evaluation. Six datasets are utilized for the experiment. Every datasets are multivariate dataset for binary classification. The number of dimension and number of instance for each class are shown in Table 1. In this paper, we call class labels as Class 1 and Class 2. Datasets are normalized based on by z-score.

We create zero-shot situation from the datasets. In zero-shot situation, training data consists of only data for train class. Also, testing data consists of data for train class and target class.

In this experiment, one class is utilized as train class. Also, another class is utilized as target class. Half of the data for train class is utilized for training. Another half of data for train class and all data for target class are utilized for testing.

**Table 1.** Description of datasets

| dataset | Class labels | dimension | Class 1 | Class 2 |
|---|---|---|---|---|
| banana | 1 or -1 | 2 | 2376 | 2924 |
| magic | g or h | 10 | 12332 | 6688 |
| phoneme | 1 or 0 | 5 | 1586 | 3818 |
| ring | 1 or 0 | 20 | 3736 | 3664 |
| twonorm | 1 or 0 | 57 | 1812 | 2785 |
| spambase | 1 or 0 | 20 | 3697 | 3703 |

## 4.2 Measurement of the evaluation

Evaluation is done using Recall. In this paper, we consider the result of train class and the result of target class separately. Hence, recall is calculated for train class and target class as given in formula (4)-(5). Also, the confusion matrix is shown in Table 2.

$$Recall_{train} = \frac{T_{train}}{T_{train}+F_{Target}} \quad (4)$$

$$Recall_{target} = \frac{T_{target}}{T_{target}+F_{Train}} \tag{5}$$

**Table 2.** Confusion Matrix

| | | Predicted | |
|---|---|---|---|
| | | Train Class | Target Class |
| Actual | Train Class | $T_{train}$ | $F_{target}$ |
| | Target Class | $F_{train}$ | $T_{target}$ |

### 4.3 Experiment result

Experiment result depends on the K-value. For each dataset, the relationship between K-value (1<=K<=200) and Recall is shown in Figure 6 to Figure 11. Each Figure consists of two graphs. For each Figure, left graph shows the result when class 1 is utilized as train class. Also, right graphs show the results when class 2 is utilized as target class. For each graph, solid line represents $Recall_{train}$ and dotted line represents $Recall_{target}$. Also, the vertical axis is Recall, the horizontal axis is K-value.

We think the best K-value should provide balanced $Recall_{train}$ and $Recall_{target}$. Hence, the best K-value is considered as where $Recall_{train}$ and $Recall_{target}$ intersect. Also, we think small K is better for scalability. Therefore, if $Recall_{train}$ and $Recall_{target}$ intersect at a small K, it is considered as good result.

On the basis of the above, we report experiment results. First, the experiments results for banana dataset and two norm dataset are shown as below.

The experiment result for banana dataset is shown in Fig. 6. Where class 1 is utilized as train-class, $recall_{train}$ and $recall_{target}$ is intersect. Based on the intersect point, the best K is 142 and the Recall is 0.75. Where class 2 is utilized as train class, $recall_{train}$ and $recall_{target}$ is intersect. Based on the intersect point, the best K is 132 and the Recall is 0.79.

Also, the experiment result for two norm dataset is shown in Fig. 7. Where class 1 is utilized as train class, $recall_{train}$ and $recall_{target}$ is intersect. Based on the intersect point, the best K is 129 and the Recall is 0.8. Where class 2 is utilized as train class, $recall_{train}$ and $recall_{target}$ is intersect. Based on the intersect point, the best K is 123 and the Recall is 0.8.

In banana dataset and twonorm dataset, the best K-values are lower than 200. It is suitable for scalability. Also, In both datasets, recalls are higher than 0.7. Hence, we can consider proposed method achieves good result for banana and twonorm.

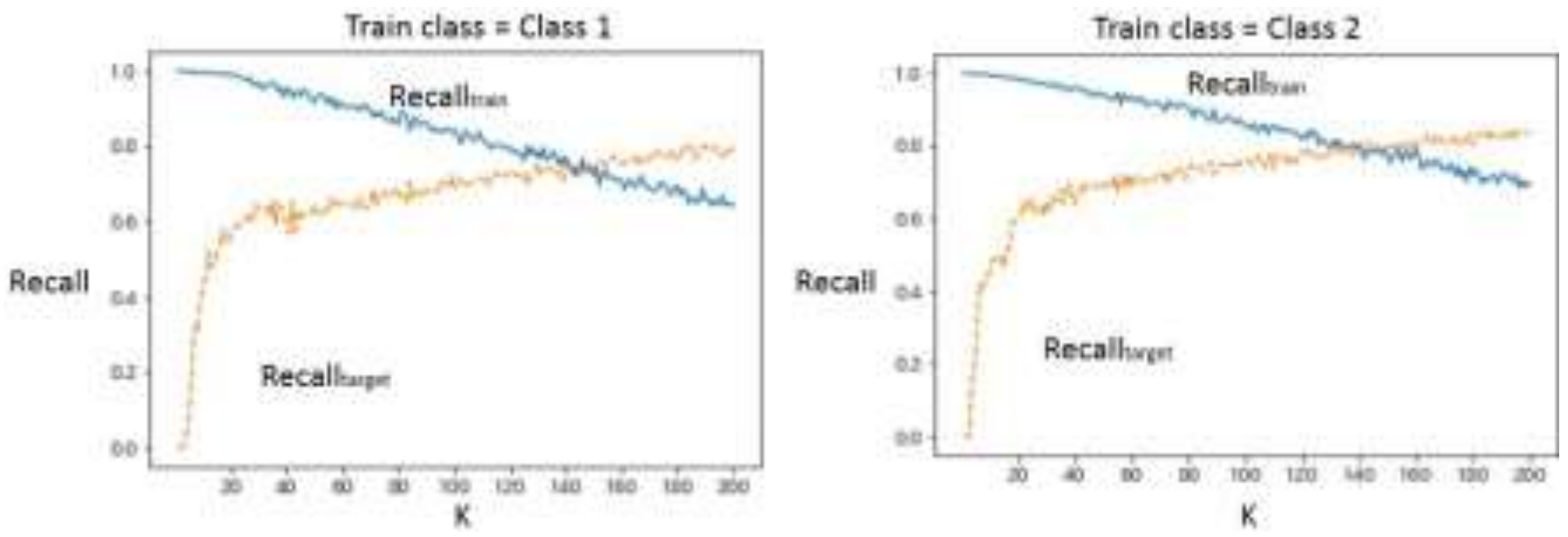

**Fig. 6.** The experiment results for banana dataset

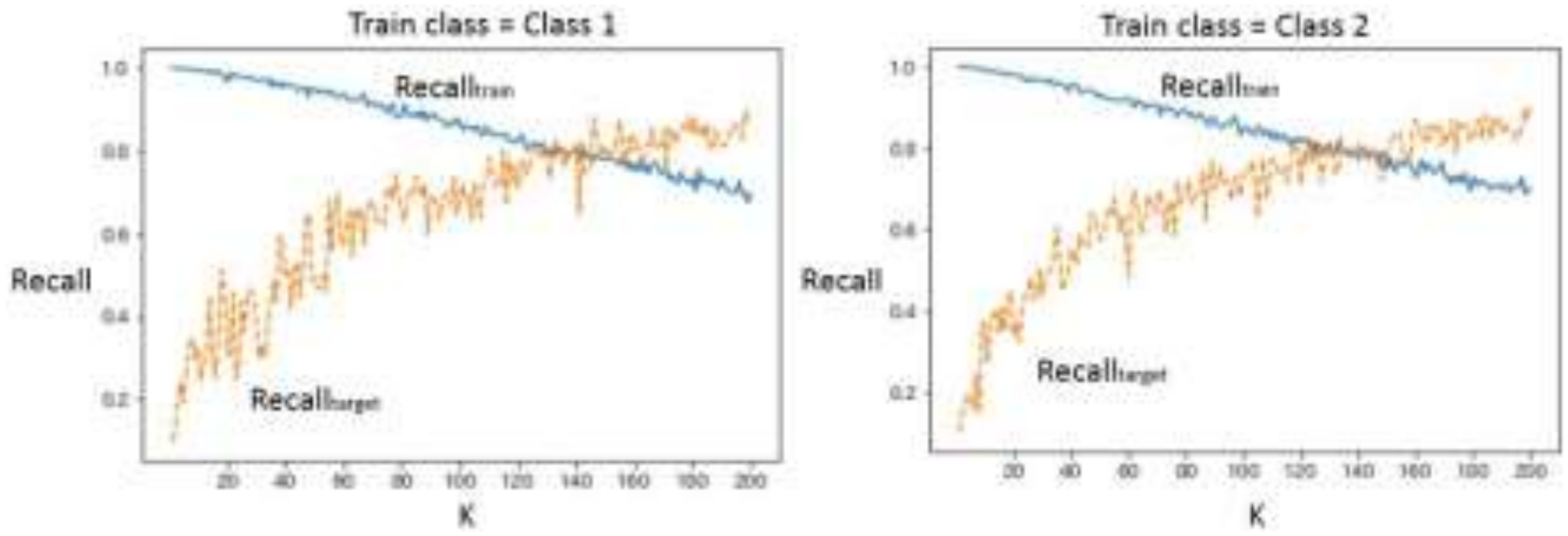

**Fig. 7.** The experiment results for two norm dataset

The experiments results for magic dataset and spambase dataset are shown as below. The experiment result for magic dataset is shown in Fig. 8. Also, the experiment result for spambase dataset is shown in Fig. 9. In Fig. 8 and Fig. 9, $recall_{train}$ and $recall_{target}$ are not intersecting in the both graphs. Hence, the best K value is higher than 200. This is not suitable in terms of scalability.

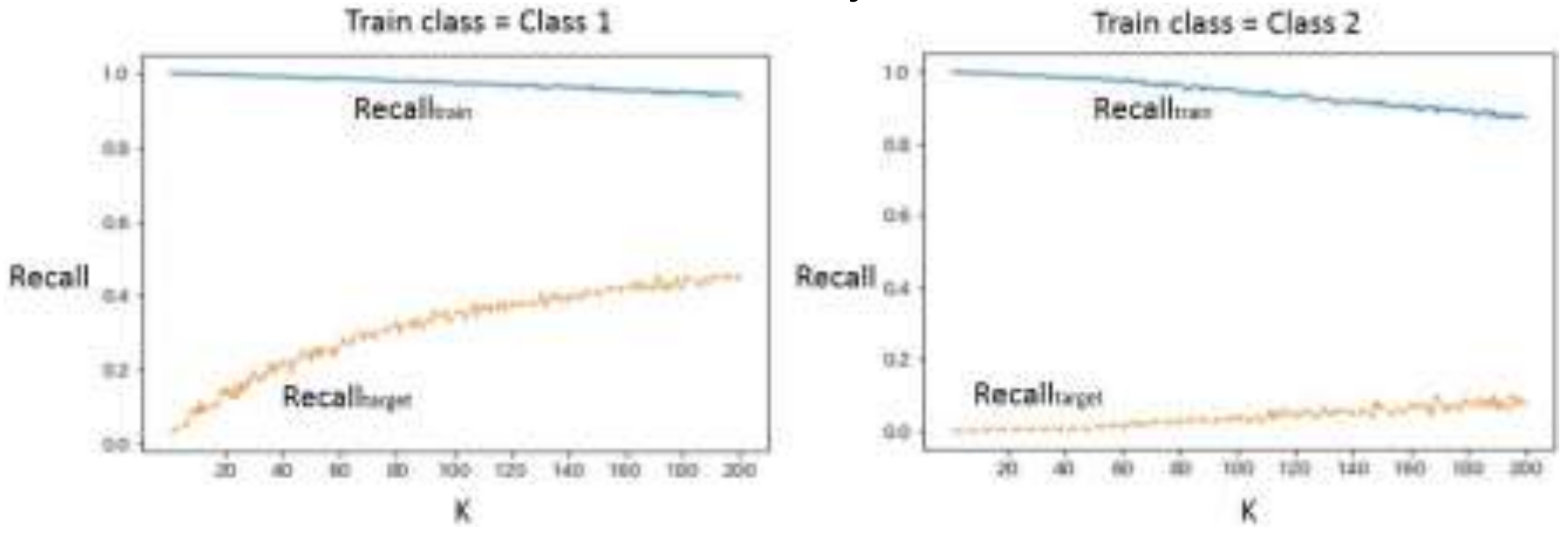

**Fig. 8.** The experiment results for magic dataset

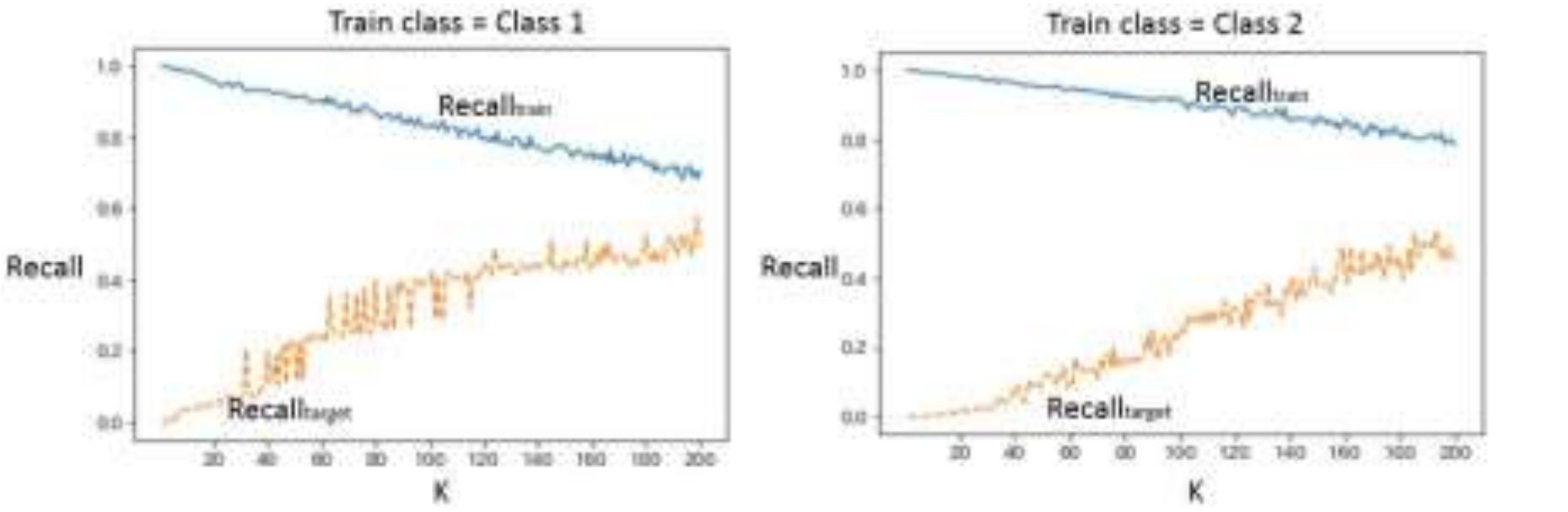

**Fig. 9.** The experiment results for spambase dataset

The experiment results for phoneme dataset and ring dataset are shown as below. The experiment result for phoneme dataset is shown in Fig. 10. Where class 1 is utilized as train class, $recall_{train}$ and $recall_{target}$ are intersecting. Based on the intersect point, the best K is 113 and the Recall is 0.73. On the other hand, where class 2 is utilized as train class, $recall_{train}$ and $recall_{target}$ are not intersecting. Hence, the best K is higher than 200.

The result for ring dataset is shown in Fig. 1. Where train class is class 1, $recall_{train}$ and $recall_{target}$ is intersect. Based on the intersect point, the best K is 27 and the Recall is 0.96. However, where train class is class 2, $recall_{train}$ and $recall_{target}$ is not intersecting. Hence, the best K value is higher than 200.

In magic dataset and ring dataset, when class 1 is utilized as train class, the best K-value is lower than 200. However, when class 2 is utilized as target class, the best K-value is higher than 200. In this case, training from one side is easy. However, training from other side is not easy. We think such problem is related to the data distribution.

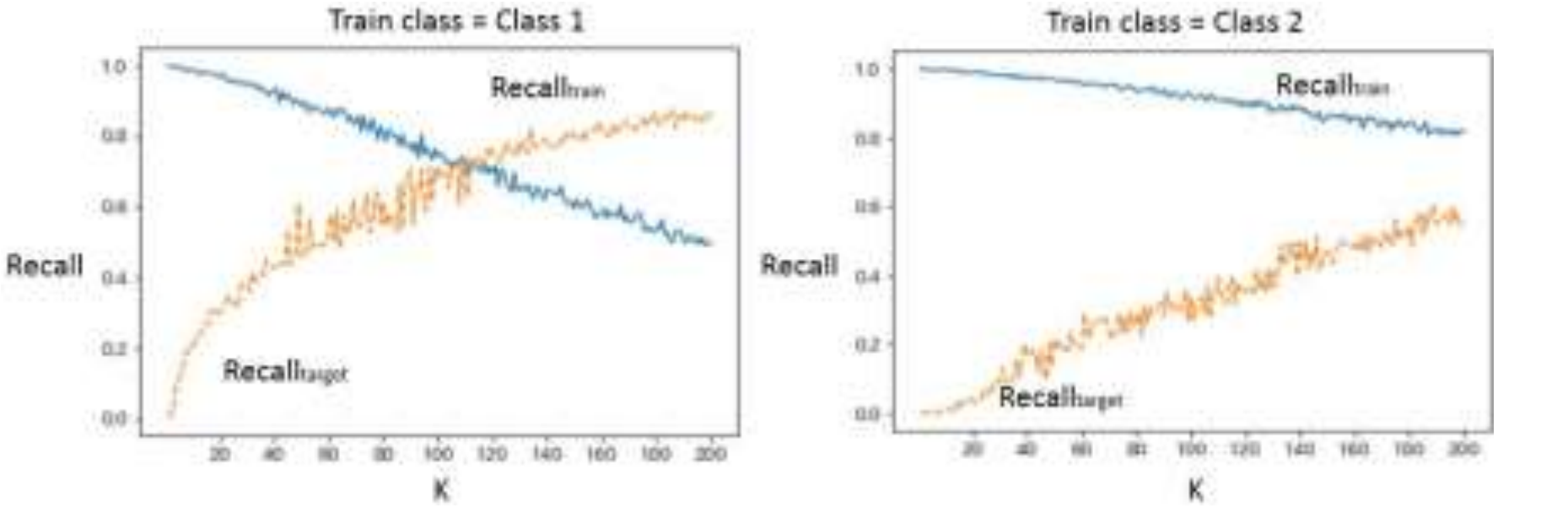

**Fig. 10.** The experiment results for phoneme dataset

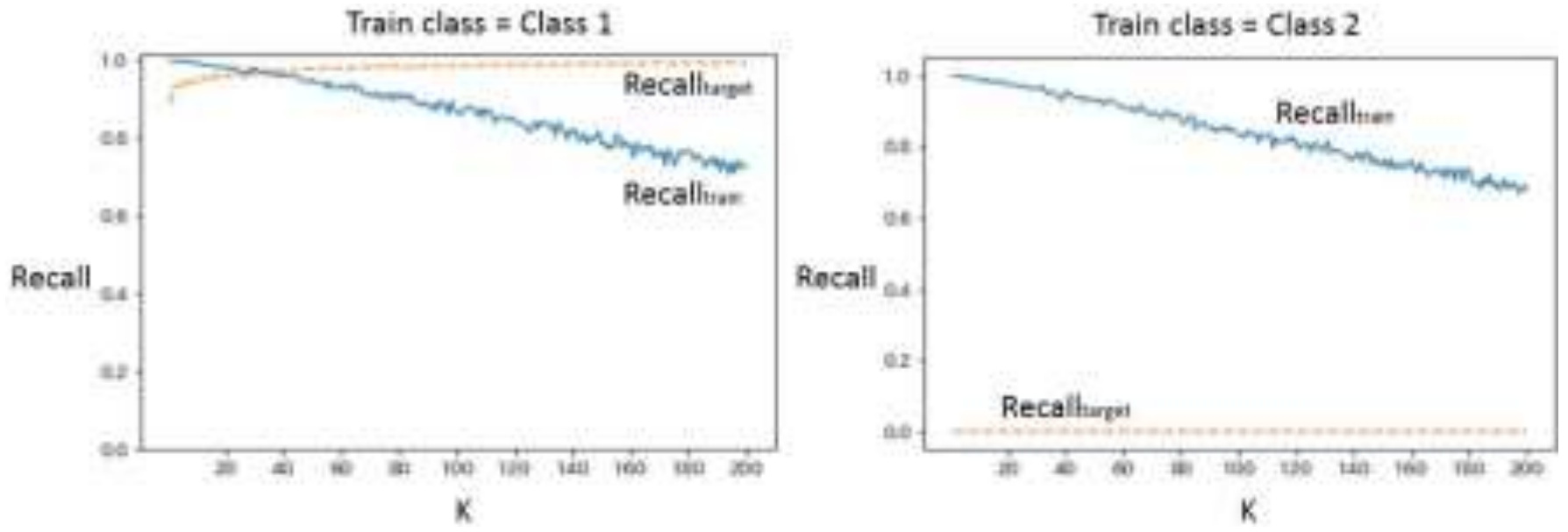

**Fig. 11.** The experiment results for ring dataset

In Figure 6 - Figure 11, Recall $_{train}$ and Recall $_{target}$ are in trade-off. Hence, selecting suitable K-value is required. We report the best (Recall$_{train}$ and Recall$_{target}$ are most balanced) results as shown in Table 3. In Table 3, the most balanced K-value and both recalls are shown with the dataset name, utilized train-class and target-class. We have investigated the best K that is higher than 200 for magic, phoneme, ring and spambase dataset. In the investigation, K is increased by 50 increments. As a result of Table 3, the best result is ring dataset where class 1 is utilized as train class. The recall is higher than 0.95. The worst result is ring dataset where class 2 is utilized as train-class The Recall is worse than 0.2. Most result is higher than 0.5. Hence, proposed method achieves good result. However, the result of magic dataset (train class = class 2) and the result of ring dataset (train class = class2) are less than 0.5. We think such problem about magic dataset is related to the class imbalance in the dataset itself. Also, we think such problem about ring dataset is related to the data distribution.
Also, smaller K-value provides better result. For example, in ring dataset (train class=class 1), K is 27 and the recall is 0.95. Also, banana and twonorm dataset has stable result. On other hand, in magic dataset (train class=class 2) and ring dataset (train class=class 2), K is 850 and the Recall is lower than 0.5. Hence, in order to improve accuracy, small K is required.

**Table 3.** The experiment results with the best K value

| dataset | Train class | Target class | K | Recall$_{train}$ | Recall$_{target}$ |
|---|---|---|---|---|---|
| banana | Class 1 | Class 2 | 142 | **0.75** | **0.75** |
| | Class 2 | Class 1 | 132 | **0.79** | **0.79** |
| magic | Class 1 | Class 2 | 700 | 0.74 | 0.73 |
| | Class 2 | Class 1 | 850 | 0.46 | 0.46 |
| phoneme | Class 1 | Class 2 | 113 | 0.72 | 0.74 |
| | Class 2 | Class 1 | 300 | 0.68 | 0.70 |
| ring | Class 1 | Class 2 | **27** | **0.95** | **0.97** |
| | Class 2 | Class 1 | 850 | 0.18 | 0.11 |
| twonorm | Class 1 | Class 2 | 129 | **0.79** | **0.82** |
| | Class 2 | Class 1 | 123 | **0.79** | **0.82** |
| spambase | Class 1 | Class 2 | 250 | 0.62 | 0.62 |
| | Class 2 | Class 1 | 350 | 0.66 | 0.68 |

# 5 Discussion

In this section, we discuss about experiment results and highlighted issues in the experiment. In table 3, the experiment results of banana, phoneme, two norm and spambase are higher than 0.6. Hence, we can consider cluster-based zero-shot learning achieves good results. However, results for magic (train-class is class 2) and ring (train-class is class 2) is lower than 0.5. This requires more preprocessing and scaling to achieve better performance.

## 5.1 The problem of ring dataset

In experiment, result of ring dataset was insufficient. If class 1 is utilized as training data, recall is higher than 0.95. However, if class 2 is utilized as training data, the recall is under 0.2. It is because problem of the distribution of the data as shown in Figure 12. In ring dataset, class 2 is surrounding class 1. Hence, training for class 1 is easy but training for class 2 is not easy. Hence this distribution should be transformed into suitable vector space.

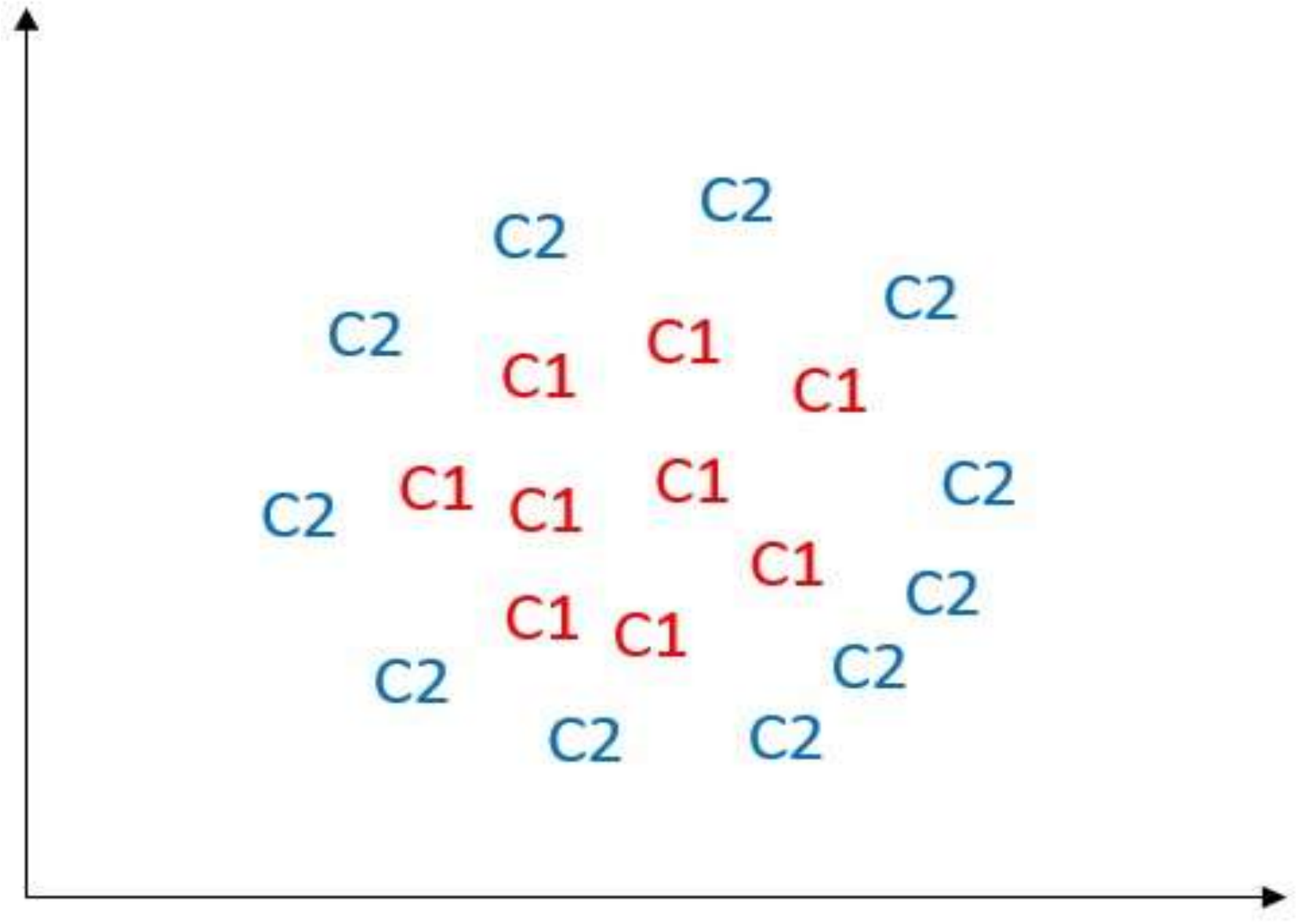

**Fig. 12.** The distribution of ring dataset

## 5.2 How to decide K-value

In experiment, the result of proposed method depends on K-value. Hence, selecting suitable K-value is required. However, in real situation, deciding the best K-value is not easy. In order to decide the best K-value, validation is required. However, ZSL cannot validate data for target class while training. It is because training data has no data for target class. Also, $Recall_{train}$ and $Recall_{target}$ is trade-off. Overall, if K-value is increase, $Recall_{target}$ is increase but $Recall_{train}$ is decrease.

### 5.3 Considering other clustering algorithm and distance

In this paper, K-means is utilized as clustering algorithm for simplicity. However, other clustering algorithm(Viegasa et al. 2018) could be comapartive to ours. Moreover, Euclidean distance is utilized as distance. However, we think it is not sufficinet for high dimensional feature. Hence, other distance measures that is suitable for high dimension feature are required. Also, feature selection or dimension reduction could be considered. In order to do feature selection or dimension reduction in ZSL, the wrapper method is not suitable. It is because we cannot validate target class while training . To find the best clustering algorithm and distance, many experiment is required. This is our future work.

## 6 Conclusions and Future work

In this paper, we consider zero-shot learning for multivariate binary classification problem. We proposed cluster-based zero-shot learning framework using K-means and Euclidean distance. This is baseline method of zero-shot learning for multivariate binary classification problem. As shown in the experiment result, result depends on K-value and $Recall_{train}$ and $Recall_{target}$ is in trade-off. Hence selecting suitable K-value is required to achive the best balance between recalls. We consider future work as follows:

- Proposed method is insufficient for some data such as ring dataset. These data should be transformed into sufficient vector space.
- The method cannot do validation while training, further solution is required.
- Trying other clustering algorithms and distances are required.
- Extending to multi-classification problem.

# Note